\documentclass[runningheads]{llncs}
\usepackage{graphicx}
\usepackage{optidef}

\usepackage{caption}
\usepackage{subcaption}

\usepackage{multirow,booktabs,color,soul}

\usepackage[ruled,linesnumbered]{algorithm2e}
\usepackage{amsmath}
\usepackage{wrapfig}

\begin{document}
\title{Multi-objective Neural Architecture Search with Almost No Training %\thanks{Supported by organization x.}
}
\author{Shengran Hu \and
Ran Cheng\thanks{Ran Cheng is the corresponding author.}  \and
Cheng He \and
Zhichao Lu
}

\authorrunning{S. Hu et al.}
\institute{
Guangdong Provincial Key Laboratory of Brain-inspired Intelligent Computation,\\
Department of Computer Science and Engineering,\\
Southern University of Science and Technology, \\
Shenzhen 518055, China.
\\
\email{
hu.shengran@outlook.com, ranchengcn@gmail.com, \\
chenghehust@gmail.com, lu.zhichao@outlook.com,
}
}
\maketitle     % typeset the header of the contribution

%
%{\color{blue} }
\begin{abstract}
In the recent past, neural architecture search (NAS) has attracted increasing attention from both academia and industries. Despite the steady stream of impressive empirical results, most existing NAS algorithms are computationally prohibitive to execute due to the costly iterations of stochastic gradient descent (SGD) training. In this work, we propose an effective alternative, dubbed \emph{Random-Weight Evaluation} (RWE), to rapidly estimate the performance of network architectures. By just training the last linear classification layer, RWE reduces the computational cost of evaluating an architecture from hours to seconds. When integrated within an evolutionary multi-objective algorithm, RWE obtains a set of efficient architectures with state-of-the-art performance on CIFAR-10 with less than two hours' searching on a single GPU card. Ablation studies on rank-order correlations and transfer learning experiments to ImageNet have further validated the effectiveness of RWE.
\end{abstract}

\section{Introduction}
% Deep Convolutional Neural Networks (CNNs) have achieved great success in various computer vision tasks. One of the key reasons behind this success is the innovation of architectures, which
% %{\color{red} From AlexNet [], VGGNet [] at the early stage, to Inception [] and ResNet [] later,}
% improves the performance of CNNs on various datasets and tasks. These early architectural advancements take not only vast computational resources but also years of efforts from researchers. The Neural Architecture Search (NAS), on the other hand, aims to alleviate this painstaking process by searching for the network architecture automatically. However, the existing NAS methods suffer from the extremely expensive computation cost. For example, the early works \cite{Zoph2018,Real2019} require thousands of GPU days to complete one search, which makes them impractical for real-world deployments.

Deep Convolutional Neural Networks (CNNs) have achieved remarkable success in various computer vision tasks. One of the key reasons behind this success is the innovation on CNN architectures \cite{Krizhevsky2012,he2016deep,muxconv}. Despite the steady stream of promising improvements over state-of-the-art across a wide range of tasks and datasets, these early architectural advancements take years of efforts from researchers and practitioners with vast computational resources. Neural Architecture Search (NAS), on the other hand, aims to alleviate this painstaking process by automating the design of CNN architectures. However, existing NAS algorithms typically require enormous computation overheads to search. For example, early works \cite{Zoph2018,Real2019} require thousands of GPU days to complete one search, making them impractical for real-world deployments.

% Early NAS methods \cite{} are primarily driven by single objective of maximizing predictive accuracy. However, real-world applications oftentimes require the CNN architectures to balance other completing objectives, such as power consumption, inference latency, memory footprint, to name a few. A number of recent method,

% The key challenge is the expensive performance evaluations for an architecture. Thorough training of a single architecture can take days to weeks on a single GPU depending on the datasets, so an estimation of performance is a must during the searching. Earlier works \cite{Zoph2018,Real2019} evaluate the networks with smaller network sizes and fewer epochs to estimate the performance under thorough training, but the acceleration is limited and the effectiveness could not be ensured in all cases \cite{zela2018}. Recent works \cite{Liu2019} adopt the weight sharing techniques to avoid redundant training, but it could not be easily adapted to various search space.

The main computation bottleneck of a NAS algorithm resides in the step of evaluating architectures' performance. A thorough training of an architecture can take days, even weeks on a single GPU card depending on the complexity of architectures and the scale of the datasets. Hereby, advocating for substitute measurements becomes a common theme in existing NAS algorithms. For instances, there are works \cite{Zoph2018,Real2019,lu2020} using the performance measured on architectures with reduced sizes and training epochs; and there are works \cite{Liu2019,lu2020nsganetv2} using performance measured on shared (instead of trained) weights. Conceptually, both of these two methods are restricted by the generality of search spaces, where the former requires search spaces to be modular and the later requires search spaces to be sequential \cite{Sandler2018,lu2020nsganetv2} (as oppose to multi-branched \cite{Zoph2018,Liu2019,lu2020}).

% In this work, we propose the random-weight evaluation, the key idea of which is to utilize the random weights after the initialization and bypass most of the training to estimate the performance. On a single Nvidia 2080Ti GPU, our method spends less than ten seconds in running an estimation for an architecture on CIFAR-10, and it is well adaptive to various NAS frameworks. Also, as shown in section \ref{section4.1}, for the last generation, the Spearman rank-order correlation of our method is higher than the one of a certain training-based method.

To address the aforementioned issues, we propose the Random-Weight Evaluation (RWE), a flexible and effective method to accelerate the performance evaluation of architectures. By leveraging the expressive power of randomly initialized convolution filters \cite{Saxe2011,lbcnn}, RWE freezes the backbone\footnote{All layers prior to the task-specific heads, e.g., the last linear layer in case of object classification.} part of CNN architectures, and only trains the last classification layer. The subsequent performance becomes the indicator to select architectures. In short, our key contributions are summarized below:

\begin{itemize}
    \item In this work, we propose RWE to expedite the performance estimation of CNN architectures. RWE is conceptually flexible as it is independent of search spaces, and empirically effective as we show later in the paper that it significantly reduces the evaluation wall-clock time from days to less than a minute. At the same time, RWE reliably estimates the performance of CNN architectures, measured in rank-order correlations.  \\

    \item We further design a multi-objective evolutionary NAS algorithm to effectively utilize the proposed evaluation method. On CIFAR-10, the proposed algorithm achieves a set of efficient architectures with state-of-the-art performance, yielding 2.98\% Top-1 error and 1.5M parameters with less than two hours on a single GPU card.
\end{itemize}

\section{Related Works}
In this section, we provide a brief overview on the topics that are closely related to technicalities of our approach. 

% \subsubsection{Performance Estimation:}
% Early works \cite{Zoph2018,Real2019} adopt the lower fidelity estimation techniques that train the architectures with smaller network sizes and fewer epochs than ones required for the deployment scene. However, these techniques are not suitable for the search space like \cite{Tan2019a,cai2020once}, in which the parameters of network sizes, like the number of layers or the channels, themselves are one of the elements to be searched. Also, a recent work  \cite{zela2018} argues that the rank-order correlation between the estimated and true performance could be low when the training resource budgets between cheap approximations and true evaluations differ too much.

\vspace{0.5em}
\noindent\textbf{Performance Estimation:}
A commonly used technique, in early NAS works \cite{Zoph2018,Real2019}, to expedite the performance evaluations involves down-scaling the architecture sizes, by reducing the number of layers, channels, and training epochs. The main limitation of this method is that it is only applicable to modular search spaces, where a CNN architecture is constructed by repeatedly stacking modular blocks. Extending this method to the search space that allows non-modular architectures is not trivial. Concurrently, there is a study \cite{zela2018} showing that extensively reducing the number of training epochs leads to a poor rank-order correlation between predicted and true performance.  

Instead of training every architecture from scratch, the \emph{weight sharing} method attempts to speed up the peformance evaluations by sharing the weights among architectures sampled during search. NAS algorithms in this category \cite{Liu2019,lu2020nsganetv2} typically construct a supernet (prior to the search), such that all searchable architectures become subsets (of the supernet; i.e. subnets) and weights are directly inherited instead of randomly initialized. Then the evaluation of an architecture becomes an inference on the validation set, which is much cheaper than training. Despite the efficiency gained during search, the supernet, required by the weight sharing method, can be more time consuming to train than a complete search \cite{cai2020once}, and may not be feasible for all search spaces, e.g. \cite{Liu2018a}.

% \subsubsection{Random weights in neural network:}

% Prior works have tried either to utilize the random weights in CNN for classification \cite{Jarrett2009} 
% %{\color{red} and other visual tasks []} 
% or to apply a single shared weight that achieves competitive performance on several RL tasks \cite{wann2019}. However, instead of utilizing the random weights directly in deployment scenes, we only take the performance with random weights as an indicator for the performance after thorough training. Similarly, the work \cite{Saxe2011} uses random weights to predict the performance ranking for some simple CNNs with few layers and the work \cite{Rosenfeld2019} fixes most of the weights to be random and trains the remained part, such that it successfully predicts the performance ranking for some famous architectures.
% Comparing to these works aim to test the idea on limited number of toy models, our approach works in modern CNN search space, achieving state-of-the-art performance, and a significant acceleration of architecture evaluations.
\vspace{0.5em}
\noindent\textbf{Expressive Power of Randomly Initialized Convolution Filters:} 
Convolution filters, even with randomly initialized weights, are surprisingly powerful in extracting meaningful feature representations from visual inputs \cite{Jarrett2009}. A number of existing works have shown that randomly initialized convolution filters can achieve comparable performance to CNNs with fully trained convolution filters on both vision and control tasks \cite{lbcnn,wann2019}. Few trials have been attempted to estimate the performance of a neural network from randomly initialized weights in the literature. More specifically, Saxe \emph{et al.} use randomly initialized weights to predict the ranking of shallow CNNs' performance \cite{Saxe2011}; Rosenfeld and Tsotsos show that the performance ranking of widely used CNN architectures
%(e.g. VGG \cite{simonyan2014very}, DenseNet \cite{Huang2017}, ResNet \cite{he2016deep}) 
can be predicted by training a tiny fractions of the weights in the CNNs \cite{Rosenfeld2019}. In this work, we train the last classification layer while freezing all other weights at initial values, and use this performance as the indicator to compare architectures. We demonstrate that our approach scales to modern CNN search spaces containing deep and complex CNNs. 

% By learning a linear combination of randomly initialized convolution filters, LBCNN achieves comparable performance to CNNs with fully trained convolution filters \cite{lbcnn}. 

\vspace{0.5em}
\noindent\textbf{Multi-objective NAS:} Early NAS algorithms \cite{Zoph2018,Liu2019,Real2019} are primarily driven by a single objective of maximizing predictive accuracy. However, real-world applications oftentimes require the CNN architectures to balance other completing objectives, such as power consumption, inference latency, memory footprint, to name a few. A portfolio of recently emerged NAS works scalarizes multiple objectives into a composite measurement that simultaneously promotes predictive performance and penalizes architecture complexity \cite{Tan2019,Cai2019}. In this work, we opt for evolutionary multi-objective optimization to approximate the entire efficient frontier in one run \cite{deb2002fast}. 

\section{Proposed Approach}
The problem of designing optimal architectures under multiple objectives for a target dataset $\mathcal{D} = $ $\{ \mathcal{D}_{trn}, \mathcal{D}_{vld}, \mathcal{D}_{tst} \}$
 can be formulated as the following bilevel optimization problem \cite{lu2020},
\newcommand{\argmin}{\operatornamewithlimits{argmin}}
\begin{mini*}|l|
{\boldsymbol{\alpha}}{f_1(\boldsymbol{\alpha};\boldsymbol{w^*}(\boldsymbol{\alpha})),f_2(\boldsymbol{\alpha}),...,f_m(\boldsymbol{\alpha})}
{}{}
\addConstraint{\boldsymbol{w^*}(\boldsymbol{\alpha}) \in \argmin_{\boldsymbol{w}} \mathcal{L}(\boldsymbol{w};\boldsymbol{\alpha})}
\addConstraint{\boldsymbol{\alpha} \in \Omega_\alpha, \hspace{1em}\boldsymbol{w} \in \Omega_w,}
% \addConstraint{\boldsymbol{w} \in \Omega_w}
\end{mini*}
where the upper lever variable $\boldsymbol{\alpha}$ defines a candidate architecture, and the lower level variable $\boldsymbol{w}(\boldsymbol{\alpha})$ represents the weights with respect to it. $\mathcal{L}(\boldsymbol{w};\boldsymbol{\alpha})$ is the cross-entropy loss on the $\mathcal{D}_{trn}$ for a candidate architecture $\boldsymbol{\alpha}$ with weights $\boldsymbol{w}$. The first objective $f_1$ represents the classification error on $ \mathcal{D}_{vld}$, which depends on both architectures and weights. Other objectives $f_2,...,f_m$ only depend on architectures, such as the number of parameters, floating-point operations (FLOPs), latency, etc. In our approach, we use predictive performance to approximate the ground-truth one and take the predictive performance and FLOPs as two objectives to be optimized, which aims to balance between the performance and the complexity of architectures.

\subsection{Random-Weight Evaluation}

\begin{algorithm}[!hbt]
    \SetAlgoLined
    \SetKwInOut{Input}{Input}\SetKwInOut{Output}{Output}
    \Input{An architecture $\boldsymbol{\alpha}$, a training and validation dataset \begin{math}\mathcal{D}_{trn}\text{,  }\mathcal{D}_{vld}\end{math}, the number of linear classifiers $L$.}
    \Output{The predictive classification error rate \emph{Error} on \begin{math}\mathcal{D}_{vld}\end{math} and the floating-point operations \emph{FLOPs} of $\boldsymbol{\alpha}$.}

    \emph{net} $\leftarrow$ Decode the architecture $\boldsymbol{\alpha}$ into CNN;

    Randomly initialize the \emph{net};

    Freeze the weights throughout the whole algorithm;

    \emph{list\_clsfr} $\leftarrow$ Randomly initialize a list of linear classifiers with length $L$;

    $\mathcal{S} \leftarrow$ Split the $\mathcal{D}_{trn}$ uniformly into $L$ subsets;

    \For{$i\leftarrow 1$ \KwTo $L$}{

        %\tcp{Train $i$-th linear classifier with a part of the $\mathcal{D}_{train}$.}

        \uIf{L $\neq$ 1}{
            $\mathcal{D}_{trn,i} \leftarrow \bigcup \mathcal{S}[j]$ for $j=1,...,L$ and $j \neq i$\;
        }
        \Else{
            $\mathcal{D}_{trn,i} \leftarrow {D}_{trn}$\;
        }

        \emph{Features} $\leftarrow$ Infer the \emph{net} on $\mathcal{D}_{trn,i}$;

        Train the linear classifier \emph{list\_clsfr}[i] with the \emph{Features} as input.

    }

    \ForEach{image, target $\in$ $\mathcal{D}_{vld}$}{
        %\tcp{Calculate the classification accuracy of \emph{net} on $\mathcal{D}_{valid}$ with ensemble linear classifiers.}

        \ForEach{linear classifier $\in$ list\_clsfr}{

        Infer the model for \emph{image} with \emph{linear classifier};

        }

        \emph{ensemble\_inference} $\leftarrow $ the label approved by most classifiers.

        Compare the \emph{ensemble\_inference} with \emph{target} and record the result;
    }

    \emph{Error} $\leftarrow$ Calculate the classification error rate;

    \emph{FLOPs} $\leftarrow$ Calculate the FLOPs of \emph{net};

    \Return{Error, FLOPs.}
 \caption{Random-Weight Evaluation}

    \label{algorithm1}

\end{algorithm}

The core of our approach, Random-Weight Evaluation (RWE), is shown in algorithm \ref{algorithm1}.
First, the architecture $\boldsymbol{\alpha}$ to be evaluated would get decoded into a CNN \emph{net} without the last classification layer, which is a linear classifier. Notice that, the number of channels and layers of the decoded architecture should be indicated in this step. Then, the weights of \emph{net} would be randomly initialized and then kept fixed throughout the whole algorithm.
We use a modified version of the \emph{Kaiming initialization} \cite{He} to initialize the \emph{net} (default setting in PyTorch).
After the initialization of \emph{net}, a list of linear classifiers \emph{list\_clsfr} would be initialized. Those classifiers get trained on the \emph{features} that are extracted by the \emph{net}, and each classifier could only be exposed to a part of the \emph{features}. Then all of the classifiers are combined as the last classification layer in the inference phrase via neural network ensemble technique \cite{Hansen1990}. More specifically, the length of the \emph{list\_clsfr} is set to five, and each classifier gets trained on $4/5$ \emph{features}. Compared with the case that only one linear classifier gets trained, a list of linear classifiers can help stabilizing RWE. Finally, the architecture is validated by taking the labels that get approved by most classifiers as a result. After calculating the classification error rate and FLOPs of the architecture, these two measurements of performance and complexity of architectures will be returned as two objectives to be optimized.

\subsection{Search Space and Encoding}

\begin{figure}[!hbt]
     \vspace{-0.5cm}
     \centering
     \begin{subfigure}[b]{\textwidth}
         \centering
         \includegraphics[width=\textwidth]{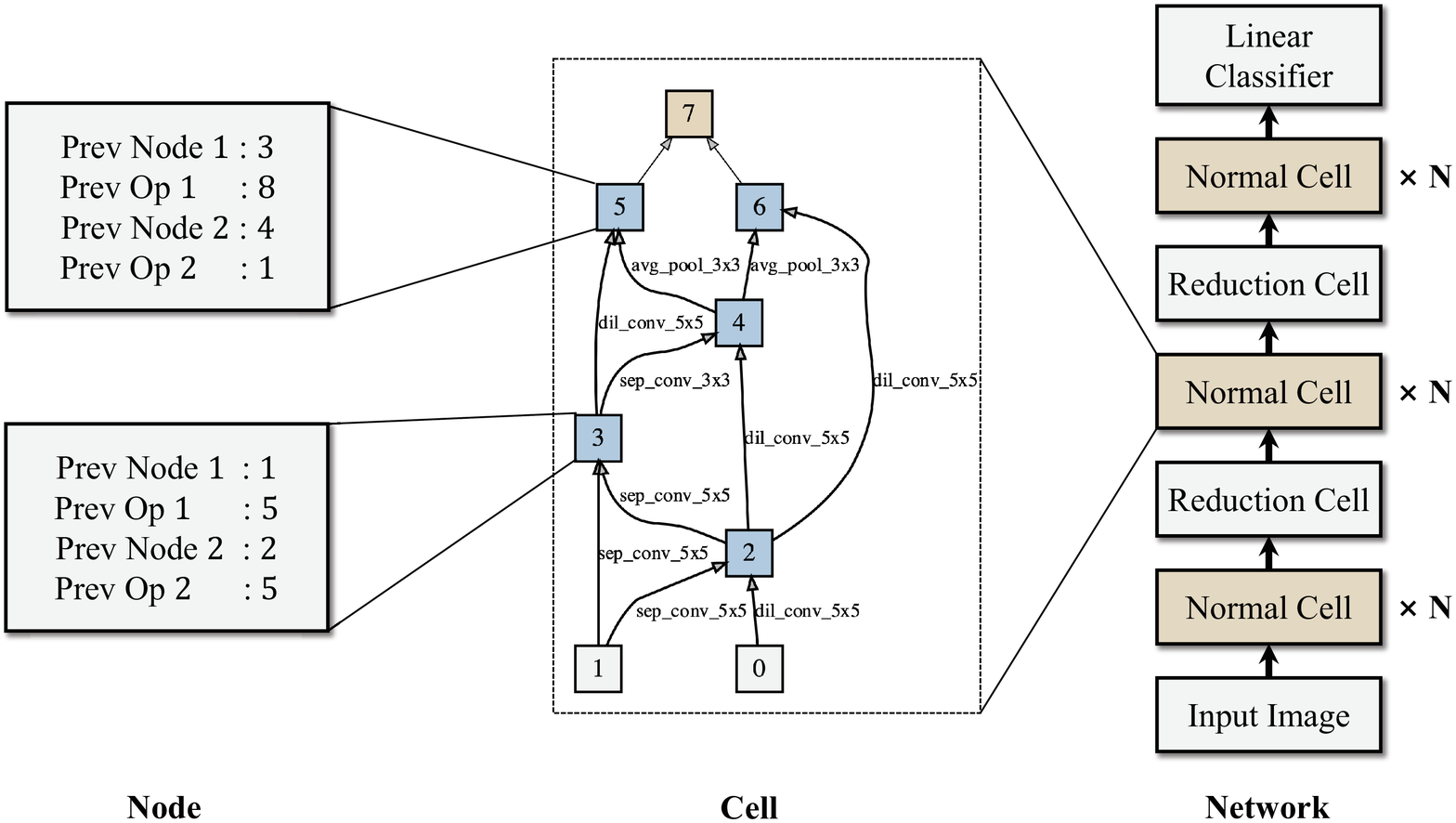}
         \caption{Search Space}
         \label{fig1a}
     \end{subfigure}
     \hfill
     \begin{subfigure}[b]{\textwidth}
         \centering
         \includegraphics[width=\textwidth]{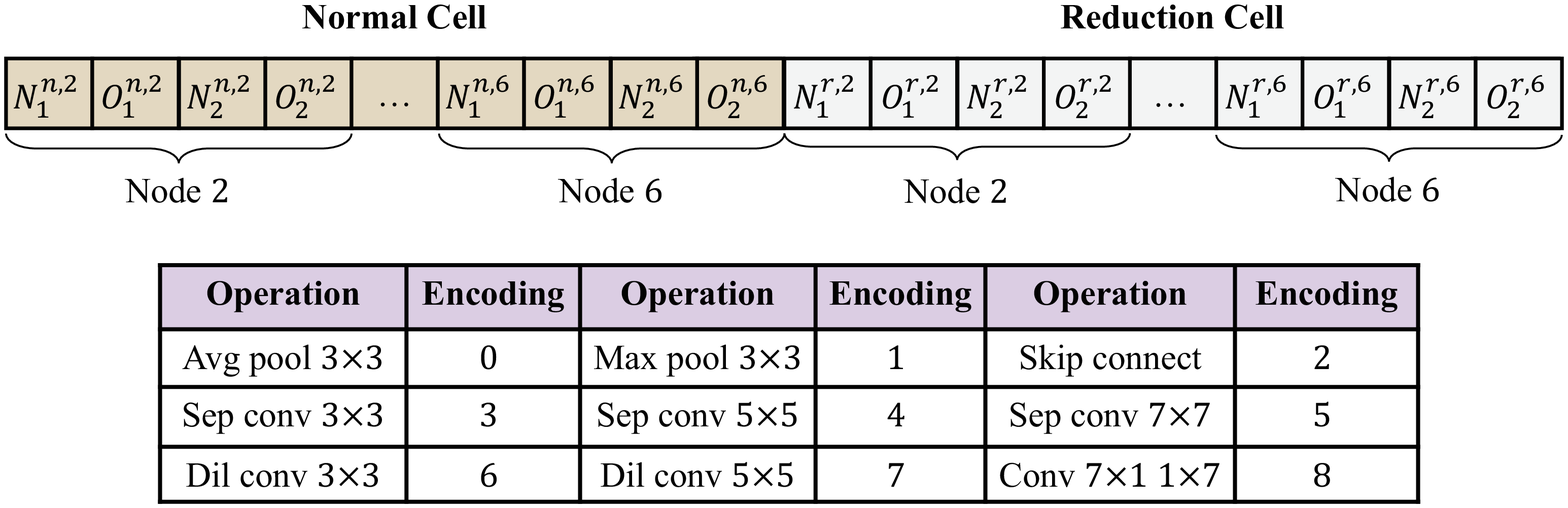}
         \caption{Encoding}
         \label{fig1b}
     \end{subfigure}
     \caption{The search space \cite{Zoph2018} adopted in our approach. (a) RIGHT: The macro network architecture of CNN. MIDDLE: An example structure of normal cells and reduction cells. LEFT: The searching elements for a node. (b) TOP: A 40-integer vector defining an architecture. BOTTOM: Candidate operations and their encoding. %The Sep Conv represents the depthwise-seperable convolution and the Dil Conv represents the dilated convolution.
     %and the Conv $7 \times 1 \text{ } 1 \times 7$  represents a $7 \times 1$ convolution followed by a $1 \times 7$ convolution.
     %BOTTOM: The candidate previous nodes for the $i$-th node.
     }
      \label{fig1}
      \vspace{-0.2cm}
\end{figure}

Our proposed RWE is conceptually flexible and can be applied to various search spaces. In this work, we adopt the NASNet \cite{Zoph2018} search space, which is widely used in various NAS algorithms \cite{Liu2019,Real2019,Lu2019}. A pictorial overview of this search space and encoding is shown in Fig. \ref{fig1}.

\vspace{0.5em}
\noindent\textbf{Network:}
% \subsubsection{Network:}
The macro network architecture is prefixed to be a stack of several cells. We search for two cells: a normal cell and a reduction cell. All normal cells in an architecture share the same structure (but different weights), which is the same case for the reduction cell. The normal cell keeps the resolution and the number of channels for input tensors, while the reduction cell downsamples the resolution and double the number of channels for input tensors. (Fig. \ref{fig1a} RIGHT)

\vspace{0.5em}
\noindent\textbf{Cell and Node:}
% \subsubsection{Cell and Node:}
The structures of both normal and reduction cells are defined by five nodes (Node $2$ to Node $6$), each of which chooses two inputs from previous nodes and two operations applied on two inputs respectively. For a node $i$, we search for two inputs from previous nodes $N_1^{cell,i}$ and $N_2^{cell,i}$ and two operations $O_1^{cell,i}$ and $O_2^{cell,i}$ applied on them, where the $cell$ could be $n$ or $r$, representing normal cells and reduction cells. For those nodes that are not chosen to become the inputs of another node, their outputs would be concatenated to the output node (Node $7$).  (Fig. \ref{fig1a} MIDDLE and LEFT)

\vspace{0.5em}
\noindent\textbf{Encoding:}
% \subsubsection{Encoding:}
We use an integer vector to encode an architecture, as shown in Fig. \ref{fig1b}. Each input and operation to be applied is encoded into an integer denoting a choice. A cell consists of five nodes, and each node is represented using four integers. The integer vectors for a normal and a reduction cell are concatenated into a $40$-integer vector. %The resulting volume of the search space is approximately equal to $3.3 \times 10^{30}$.

\subsection{Search Strategy}
In our approach, we adopt the classic multi-objective evolutionary algorithm NSGA-II \cite{deb2002fast} as the searching framework. The search process is briefly summarized below.

First, the population is randomly initialized. After individuals in it get evaluated by RWE, the binary tournament selection will be applied to select parents for offspring. Then through two-point crossover and polynomial mutation, offspring is created. Finally, the offspring gets evaluated, followed by the environment selection through the nondominated sorting and the crowding distance. The process is repeated
%from the binary tournament selection step
until reaching the max generation.

\section{Experimental Results}
In this section, we first evaluate the effectiveness of RWE, followed by the results of our approach searching on CIFAR-10 \cite{krizhevsky2009learning} and transferring to ImageNet \cite{imagenet_cvpr09}.

\subsection{Effectiveness of Random-Weight Evaluation}
\label{section4.1}
\begin{figure}
    \vspace{-0.6cm}
    \centering
    \includegraphics[width=0.9\textwidth]{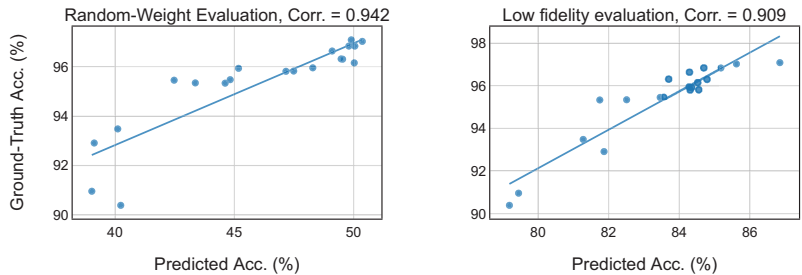}
    \caption{The predicted accuracy and ground-truth accuracy on CIFAR-10 of the architectures from final generation. The line represents a linear regression. Corr. is the Spearman rank-order correlation coefficient between the predicted and the ground-truth accuracy.
    %LEFT: Prediction from our method, random-weight evaluation. Right: Prediction from a common method, low fidelity evaluation.
    }
    \label{fig2}
    \vspace{-0.5cm}
\end{figure}

To demonstrate the effectiveness of RWE, we calculate the Spearman rank-order correlation coefficient between the predicted and the ground-truth accuracy on CIFAR-10, which is the larger the better, equal to 1.0 if the predicted rank is the same as the ground-truth one. The result is shown in Fig.\ref{fig2}. The ground-truth accuracy is obtained through the setting described in section \ref{section4.2}. Fig.\ref{fig2} LEFT shows the predictions by RWE, as described in algorithm 1.
%The correlation coefficient is equal to 0.942.
Fig.\ref{fig2} RIGHT shows the predictions by low fidelity evaluation commonly adopted \cite{Zoph2018,Real2019,Lu2019}. These predictions are obtained by a twenty-epoch training on the architecture with 8 layers and 16 initial channels.
%The correlation coefficient is equal to 0.909.
The correlation coefficients for these two methods are 0.942 and 0.909, respectively.
The result shows that, in this circumstance, RWE predicts the accuracy slightly better than the low fidelity evaluation.

\subsection{Searching on CIFAR-10\label{section4.2}}
In this work, we search on CIFAR-10, a dataset that is widely used for benchmarking image classification. CIFAR-10 is a 10-category dataset, consisting of 60K images with resolution of $32 \times 32$. The dataset is split into a training set and a testing set which consists of 50K and 10K images, respectively. Furthermore, we split the training set ($80\%-20\%$) to create the training and validation set for the searching stage.

\begin{figure}[!hbt]
     \vspace{-0.6cm}
     \centering
     \begin{subfigure}[b]{\textwidth}
         \centering
        \includegraphics[width=.80\textwidth]{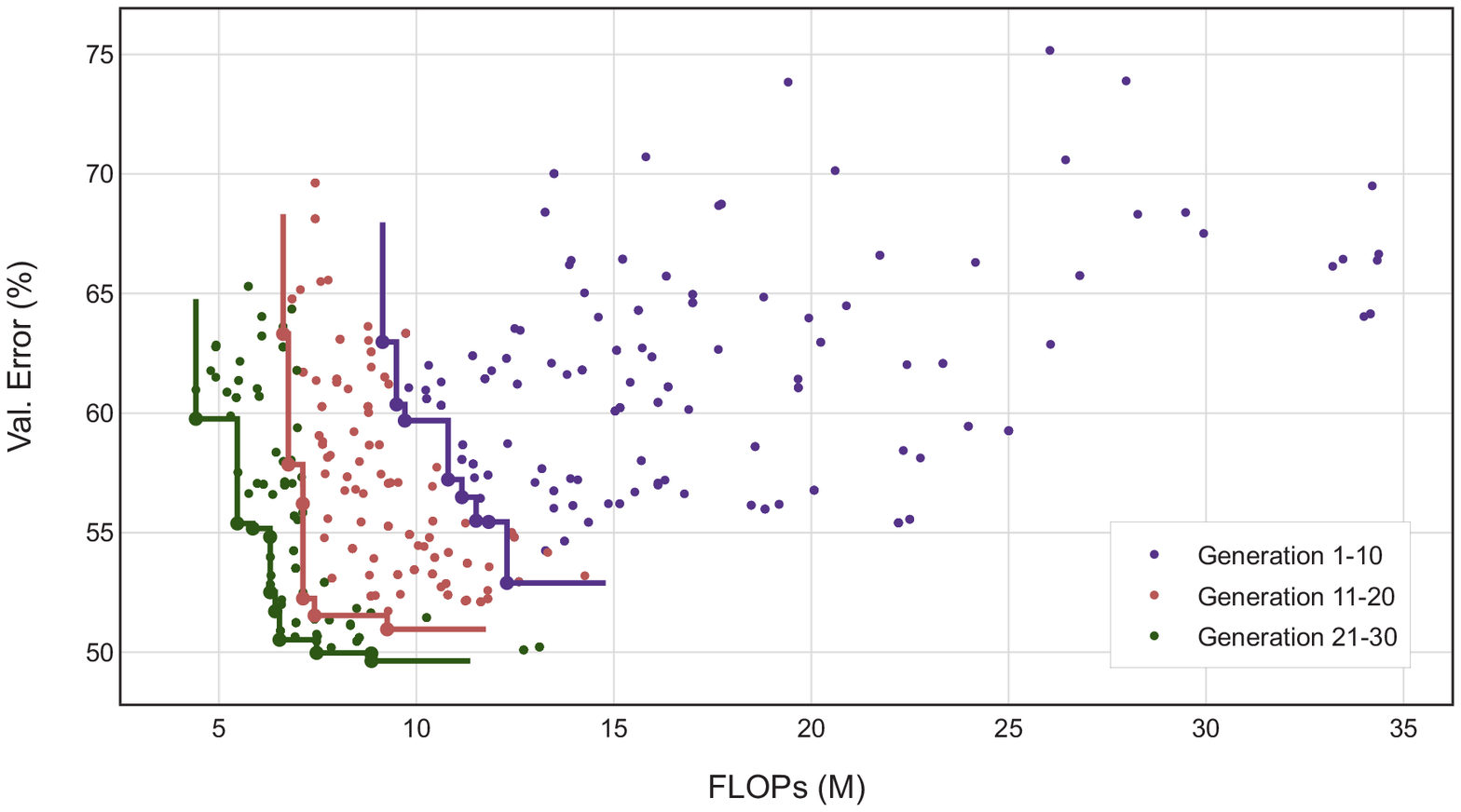}
        % \caption{Progression of Pareto fronts for different generations}
        % \label{PF}
     \end{subfigure}
     \begin{subfigure}[b]{\textwidth}
         \centering
        \includegraphics[width=.80\textwidth]{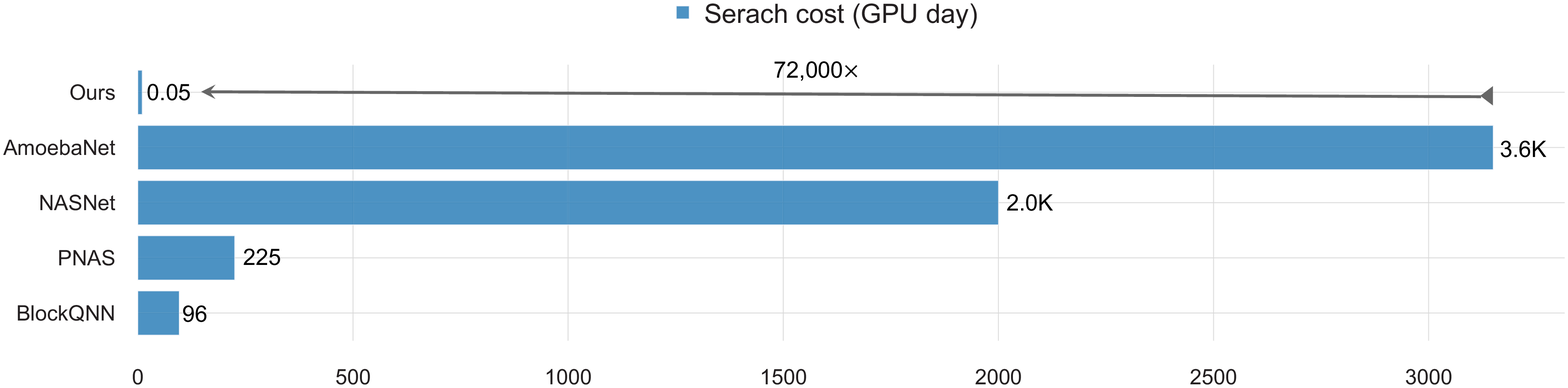}
        % \caption{Searching cost comparison}
        % \label{searchcost}
     \end{subfigure}
     \caption{TOP: Progression of Pareto fronts for different generations. BOTTOM: Search cost comparison.}
    %  \caption{The Pareto fronts from searching progress and the search cost compared to other NAS methods.}
     \label{searching}

\end{figure}

In our searching stage, the population size $N$ is set to 20 and we search for 30 generations. For RWE, every architecture is decoded into a CNN with 10 channels and 5 layers, where the second and fourth layers are reduction cells, and others are normal cells. The length of the linear classifiers list $L = 5$. The preprocessing for input images only contains the normalization, without the standard data augment techniques that introduce the randomness. The training for the linear classifiers uses the SGD optimizer with a batch size set of 512 and the momentum of 0.9. Also, the learning rate is set to 0.25 initially and then decay to zero by the cosine annealing schedule \cite{loshchilov2016sgdr}. We train each linear classifier for 30 epochs, and the search takes one hour and fifteen minutes with a single Nvidia 2080Ti. Fig.~\ref{searching} TOP shows the bi-objective Pareto fronts for different generations in the searching stage. Fig.~\ref{searching} BOTTOM shows that our approach saves orders of magnitude search cost compared with other NAS algorithms.

\begin{table}[!hbt]
\centering
\vspace{-0.5cm}
\caption{Comparisons with other state-of-the-art architectures on CIFAR-10. $^{\Updownarrow}$ denotes the work that shares the same setting with ours and the performance are reported in  \cite{Lu2019}. $^{\dagger}$ denotes the work that adopts the \emph{cutout} \cite{devries2017improved} technique.}
\resizebox{0.98\textwidth}{!}{%
\begin{tabular}{@{\hspace{2mm}}lccccc@{\hspace{2mm}}}
\toprule
Architecture &
  \multicolumn{1}{c}{\begin{tabular}[c]{@{\hspace{2mm}}c@{}}Test Error\\ (\%)\end{tabular}} &
  \multicolumn{1}{c}{\begin{tabular}[c]{@{\hspace{2mm}}c@{}}Params\\ (M)\end{tabular}} &
  \multicolumn{1}{c}{\begin{tabular}[c]{@{\hspace{2mm}}c@{}}FLOPs\\ (M)\end{tabular}} &
  \multicolumn{1}{c}{\begin{tabular}[c]{@{\hspace{2mm}}c@{}}Search Cost\\ (GPU days)\end{tabular}} &
  \hspace{2mm} Search Method \\ \hline
Wide ResNet \cite{Zagoruyko2016}       & 4.17 & 36.5 & -   & -     & manual         \\
DenseNet-BC \cite{Huang2017}             & 3.47 & 25.6 & -   & -     & manual         \\ \midrule
%PNAS     \cite{liu2018progressive}                & 3.41 & 3.2  & -   & 225   & SMBO           \\
BlockQNN$^{\dagger}$  \cite{Zhong2020}                    & 3.54 & 39.8 & -   & 96    & RL \\
SNAS$^{\dagger}$ \cite{Xie2019}    & 3.10 & 2.3  & -   & 1.5   & gradient  \\
NASNet-A$^{\dagger\Updownarrow}$  \cite{Zoph2018}                 & 2.91 & 3.2  & 532 & 2,000 & RL             \\
%ENAS$^{\dagger\Updownarrow}$  \cite{Pham2018}                      & 2.89 & 3.3  & 533 & 0.5   & RL             \\
DARTS$^{\dagger\Updownarrow}$  \cite{Liu2019}         & 2.76 & 3.3  & 547 & 4     & gradient \\

%MetaQNN  \cite{Baker2019}                             & 6.92 & 11.8 & -   & 100   & RL             \\
          \midrule
 NSGA-Net$^{\Updownarrow}$ + macro space  \cite{Lu2019}                 & 3.85 & 3.3  & 1290 & 8     & evolution      \\
\textbf{Ours$^{\dagger}$ + macro space} & $\boldsymbol{4.27}$ & $\boldsymbol{2.79}$ & $\boldsymbol{1074}$ & $\boldsymbol{0.14}$  & evolution\\
 \midrule
AE-CNN + E2EPP \cite{8744404} & 5.30 & 4.3  & - & 7     & evolution      \\
Hier. Evolution \cite{Liu2018a}               & 3.75 & 15.7 & -   & 300   & evolution     \\
AmoebaNet-A$^{\dagger\Updownarrow}$  \cite{Real2019}              & 2.77 & 3.3  & 533 & 3,150 & evolution      \\
NSGA-Net$^{\dagger\Updownarrow}$  \cite{Lu2019}                 & 2.75 & 3.3  & 535 & 4     & evolution      \\

%Large-Scale Evolution  \cite{Real2017}              & 5.4  & 5.4  & -   & 2600  & evolution      \\
 \midrule
\textbf{Ours-s$^{\dagger}$} & $\boldsymbol{4.05}$ & $\boldsymbol{0.9}$ & $\boldsymbol{203}$ & $\boldsymbol{0.05}$  & evolution\\
\textbf{Ours-m$^{\dagger}$} & $\boldsymbol{3.37}$ & $\boldsymbol{1.2}$ & $\boldsymbol{249}$ & $\boldsymbol{0.05}$  & evolution\\
\textbf{Ours-l$^{\dagger}$} & $\boldsymbol{2.98}$ & $\boldsymbol{1.5}$ & $\boldsymbol{340}$ & $\boldsymbol{0.05}$  & evolution\\

\bottomrule
\end{tabular}}
\label{table1}
%\vspace{-0.5cm}
\end{table}

\begin{figure}
    %\vspace{-0.3cm}
    \centering
    \includegraphics[width=\textwidth]{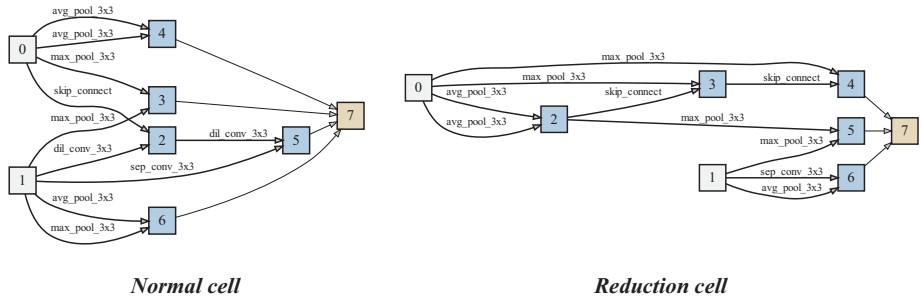}
    \caption{The visualization of \textbf{Ours-l} architecture.
    }
    \label{visualization}
    \vspace{-0.5cm}
\end{figure}

For validations, we adopt the same training setting in \cite{Lu2019} for a fair comparison. We train the architectures selected from the final Pareto Front with 20 layers and 34 initial channels from scratch. The number of epochs is set to 600 with a batch-size of 96. We also use the data augmentation technique \emph{cutout} \cite{devries2017improved} with a length of 16, and the regularization technique scheduled path dropout introduced in \cite{Zoph2018} with a dropout rate 0.2.
%We train the architectures selected from the final Pareto Front with 20 layers and 34 initial channels from scratch. The SGD optimizer is used with a learning rate of 0.025, the momentum of 0.9, and the weight decay of $3\times 10^{-4}$. The learning rate decays to zero by the cosine annealing schedule. We set the \# of epochs to 600 with a batch-size of 96 and we also use the data augmentation technique \emph{cutout} \cite{devries2017improved} with a length of 16, and the regularization technique scheduled path dropout introduced in \cite{Zoph2018} with a dropout rate 0.2. Furthermore, an auxiliary head classifier is appended right after the second reduction cell, which is aggregated with the normal loss after scaled by a constant 0.4.

The validation results and comparisons to other state-of-the-art architectures on CIFAR-10 are shown in Table \ref{table1}. We select three representative architectures from the final generation to compare with other hand-crafted and search-generated architectures. The chosen architecture with the lowest error rate (\textbf{Ours-l}, as shown in Fig. \ref{visualization}) results in a 2.98\% classification error and 340M FLOPs, which is competitive to other works in error rate but has fewer FLOPs. Also, compared with other NAS algorithms, our approach has much less computational cost measured in GPU days.

To further validate the effectiveness of RWE, we also apply our NAS algorithm on the macro search space \cite{Xie2017} adopted in \cite{Lu2019}. The results in Table \ref{table1} show that our chosen architecture has competitive performance but fewer FLOPs, similar to the case in micro search space.

\subsection{Transferring to ImageNet}

\begin{wraptable}{r}{8cm}
\vspace{-0.5cm}
\caption{Comparisons with other state-of-the-art architectures on ImageNet. $^{\Uparrow}$ denotes the architectures that are searched on CIFAR-10 and transferred to ImageNet.}
	\centering
	\begin{tabular}{lcccc}
\toprule
\multicolumn{1}{l}{\multirow{2}{*}{Architecture}} &
  \multicolumn{2}{c}{\begin{tabular}[c]{@{}c@{}}Test Error (\%)\end{tabular}} &
  \multicolumn{1}{c}{\multirow{2}{*}{\begin{tabular}[c]{@{}c@{}}Params\\ (M)\end{tabular}}} &
  \multicolumn{1}{c}{\multirow{2}{*}{\begin{tabular}[c]{@{}c@{}}FLOPs\\ (M)\end{tabular}}} \\ \cline{2-3}
\multicolumn{1}{l}{}                          & top-1 & top-5 & \multicolumn{1}{c}{} & \multicolumn{1}{c}{} \\ \midrule
MobileNetV1 \cite{Howard2017}                          & 31.6  & -     & 2.6                  & 325                  \\
InceptionV1 \cite{Szegedy2015}                                   & 30.2  & 10.1  & 6.6                  & 1448                 \\
ShuffleNetV1 \cite{Zhang2018}                          & 28.5  & -     & 3.4                  & 292                  \\
MobileNetV2 \cite{Sandler2018}                          & 28.0  & 9.0   & 3.4                  & 300                  \\

\midrule
NASNet-C $^{\Uparrow}$ \cite{Zoph2018} & 27.5  & 9.0   & 4.9                  & 558                  \\
SNAS $^{\Uparrow}$ \cite{Xie2019}          & 27.3  & 9.2   & 4.3                  & 533                  \\ \midrule
\textbf{Ours-l} $^{\Uparrow}$                  & $\boldsymbol{27.6}$  & $\boldsymbol{9.4}$   & $\boldsymbol{3.7}$                  & $\boldsymbol{363}$ \\\bottomrule
\end{tabular}

\vspace{-0.5cm}
\label{table2}
\end{wraptable}

 It is a common approach that architectures get searched on CIFAR-10 and then get transferred to other datasets or tasks \cite{Zoph2018,Xie2019}. To test the transferability of our algorithm, we transfer our architecture with lowest error rate to ImageNet \cite{imagenet_cvpr09} dataset, which is one of the most challenging datasets for image classification. It is consisted of 1.28M images for the training set and 50K images for the validation set. The images are of various resolution and unevenly distributed in 1000 categories. We adopt some common data augmentation techniques, including the random resize and crop, the random horizontal flip, and the color jitter. We adjust our network architecture for ImageNet based on \cite{Liu2019}.
 %More specifically, the architecture starts with three stem convolutional layers with stride 2, which downsample the resolution by eight times. Following, there are 14 layers, with initial channels of 48, where the reduction cells appear on the fifth and ninth layer.
 We train the model on 4 Nvidia Tesla V100 with the SGD optimizer for 250 epochs, with the batch size 1024 and resolution $224 \times 224$. The learning rate is set to 0.5, with the momentum 0.9 and the weight decay $3\times10^{-5}$. The linear learning rate scheduler is used and as a result, the learning rate decays from 0.5 to $1 \times 10 ^{-5}$ linearly during the training. Also, the warm-up strategy is adopted to increase the learning rate from 0 to 0.5 over the first five epochs. The label smooth \cite{szegedy2016rethinking} technique is also used with a smooth ratio of 0.1.
The results of comparisons of our approach to other state-of-the-art architectures on ImageNet is shown in Table \ref{table2}.

\section{Conclusion}

% In this paper, we have presented a multi-objective NAS algorithm that achieves state-of-the-art performance while having a much less computational cost. Within this algorithm, a novel performance estimation strategy, Random-Weight Evaluation (RWE), has been proposed, which drastically reduces the search cost of our algorithm. RWE ﬁxes the majority of the weights at randomly initialized values, and only trains the last linear classiﬁer layer. It costs our algorithm less than two hours (0.05 GPU days) to finish the search on CIFAR-10 and achieves an error rate of 2.98\% on it. It also achieves a 27.6\% Top-1 error rate on the ImageNet dataset after transferring the architecture searched on CIFAR-10.

In this paper, we proposed a novel performance estimation strategy, Random-Weight Evaluation (RWE), to reduce the search cost of a NAS algorithm. RWE ﬁxes majority of the weights at randomly initialized values, and only trains the last linear classiﬁer layer. We integrated RWE in a multi-objective NAS algorithm, achieving state-of-the-art performance while reducing the computational cost drastically. In particular, RWE obtained a novel architecture on CIFAR-10, yielding 2.98\% top-1 classification error and 1.5M parameters with less than two hours searching on a single GPU card. When transferred to ImageNet, the obtained architecture achieved 27.6\% top-1 classification error.

\section*{Acknowledgement}
This work was supported by the National Natural Science Foundation of China (Grant No. 61903178 and 61906081), the Program for Guangdong Introducing Innovative and Enterpreneurial Teams (Grant No. 2017ZT07X386), the Shenzhen Peacock Plan (Grant No. KQTD2016112514355531), and the Program for University Key Laboratory of Guangdong Province (Grant No. 2017KSYS008).

\bibliographystyle{splncs04}
\bibliography{wann}

%\end{thebibliography}
\end{document}